\documentclass[conference]{IEEEtran}
\IEEEoverridecommandlockouts
\usepackage{cite}
\usepackage{amsmath,amssymb,amsfonts}
\usepackage{algorithmic}
\usepackage{graphicx}
\usepackage{textcomp}
\usepackage{xcolor}
\usepackage{bm}
\usepackage{gensymb}
\usepackage{subcaption}
\usepackage{hyperref}
\def\BibTeX{{\rm B\kern-.05em{\sc i\kern-.025em b}\kern-.08em
    T\kern-.1667em\lower.7ex\hbox{E}\kern-.125emX}}
\begin{document}

\title{LFTag: A Scalable Visual Fiducial System with Low Spatial Frequency}
\author{\IEEEauthorblockN{Ben Wang}
\IEEEauthorblockA{
\textit{University of Pennsylvania}\\
bendwang@seas.upenn.edu}
}
\maketitle

\begin{abstract}
Visual fiducial systems are a key component of many robotics and AR/VR applications for 6-DOF monocular relative pose estimation and target identification. This paper presents LFTag, a visual fiducial system based on topological detection and relative position data encoding which optimizes data density within spatial frequency constraints. The marker is constructed to resolve rotational ambiguity, which combined with the robust geometric and topological false positive rejection, allows all marker bits to be used for data.

When compared to existing state-of-the-art square binary markers (AprilTag) and topological markers (TopoTag) in simulation, the proposed fiducial system (LFTag) offers significant advances in dictionary size and range. LFTag 3x3 achieves 546 times the dictionary size of AprilTag 25h9 and LFTag 4x4 achieves 126 thousand times the dictionary size of AprilTag 41h12 while simultaneously achieving longer detection range. LFTag 3x3 also achieves more than twice the detection range of TopoTag 4x4 at the same dictionary size.
\end{abstract}


\section{Introduction}
Visual fiducual systems consist of a set of markers located in the environment which can be uniquely identified and localized by an algorithm. These systems are often used in virtual/augmented reality as well as robotics applications, as many fiducial systems enable accurate 6-DOF monocular pose estimation.

There also exists many active visual fiducial systems, where the marker is powered\cite{1544677}, or require more hardware in addition to a camera to localize. These systems could include LIDAR\cite{huang2019lidartag}, or an light source at the camera in addition to retroreflective markers\cite{10.1145/1531326.1531404}. These systems are not discussed in this paper, as the focus is purely on passive visual fiducial systems with markers which can be cheaply produced on a standard printer.

The challenges of designing a visual fiducial systems is three fold: the amount of unique markers should be high, but simultaneously the markers must be have a low false positive rate and markers must be computationally efficient to detect.

\begin{figure}[htbp]
\centerline{\includegraphics[width=1.0\linewidth]{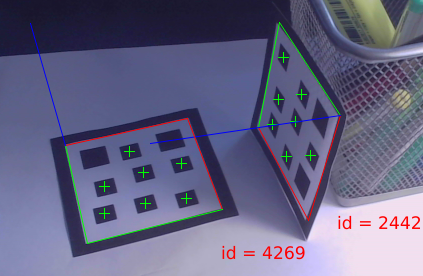}}
\caption{Example of two LFTags in the real world, detected, decoded and localized in challenging lighting and perspective conditions}
\label{lftagexample}
\end{figure}

This paper presents a novel visual fiducial system (see example in fig~\ref{lftagexample}) with high data density while being detectable from a long range. The proposed visual fiducial is tested against state of the art systems in simulation.

\section{Previous Work}

There have been many previous visual fiducial systems, with a multitude of different approaches to marker detection and data encoding. The three general approaches are dense binary markers, relative positioning based markers, and topology based markers, which are described in the following sections.

\subsection{Square Binary Markers}

Square binary markers encode data with a binary grid where each square is either black or white, usually delineated from the background with a solid black border.

ARTag\cite{artag} uses this structure, with an interior binary grid which varies between 4x4 and 6x6. Detection is done with an image gradient based quad detector. A coding and forward error correction scheme was created which minimized marker cross correlation while correcting for up to two bit errors.

AprilTag\cite{apriltag} improved the coding scheme with a lexicode construction, which maximized the minimum hamming distance within the dictionary. Various minimum complexity metrics were also employed for the binary pattern within the marker\cite{apriltag3}, which reduced false positives by eliminating likely patterns in the wild, at some cost of dictionary size.

\subsection{Relative Positioning Based Markers}
\begin{figure}[htbp]
\centering
\begin{subfigure}[b]{0.4\linewidth}
\includegraphics[width=1\linewidth]{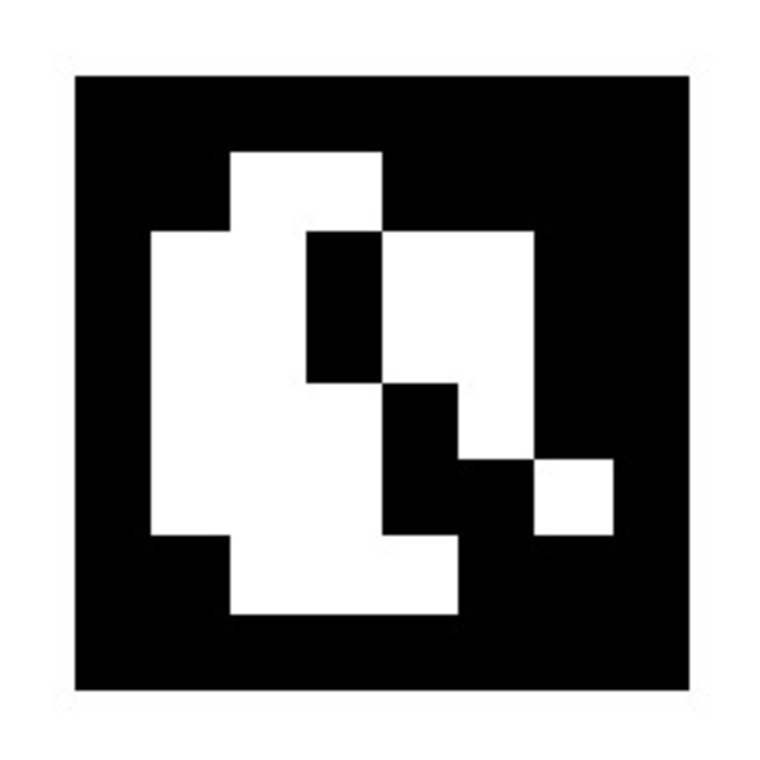}
\caption{ARTag 6x6}
\label{artag}
\end{subfigure}
\begin{subfigure}[b]{0.4\linewidth}
\includegraphics[width=1\linewidth]{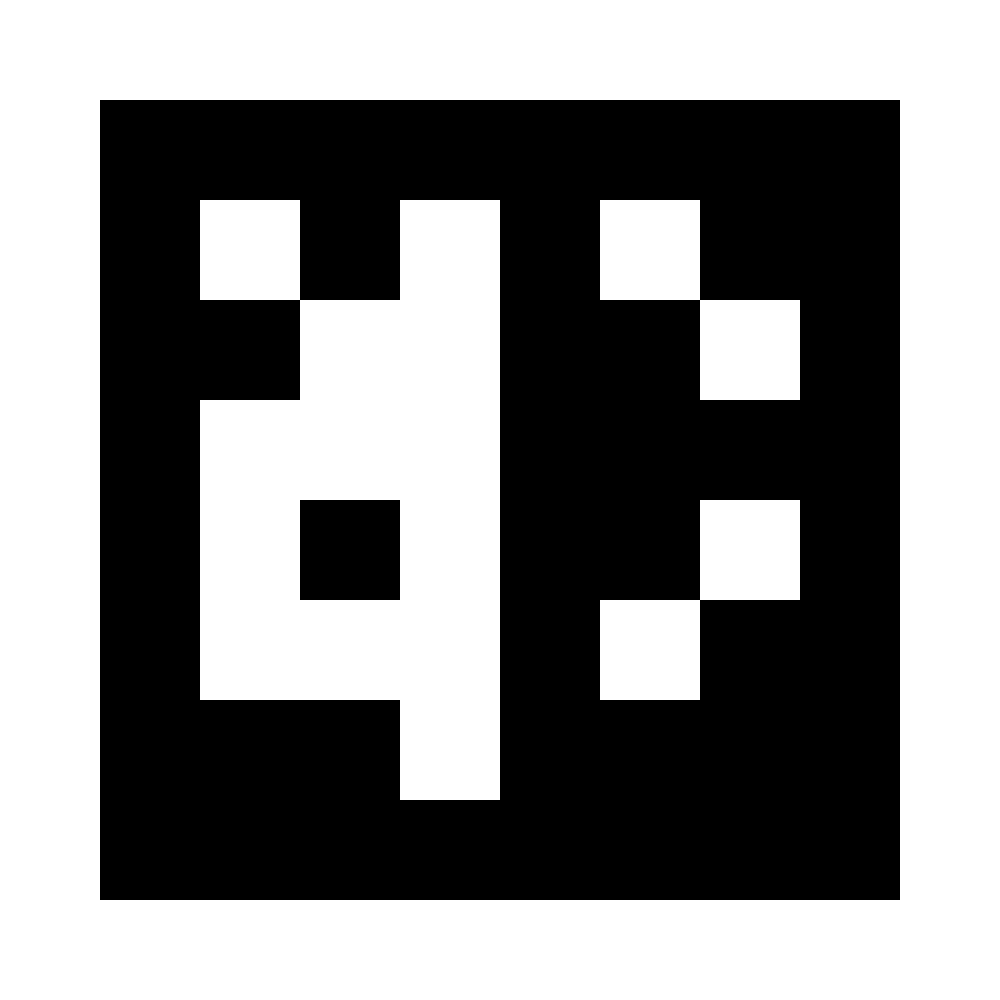}
\caption{AprilTag 36h11}
\label{apriltag}
\end{subfigure}
\caption{Two examples of square binary markers.}
\end{figure}

\begin{figure}[htbp]
\centering
\includegraphics[width=0.4\linewidth]{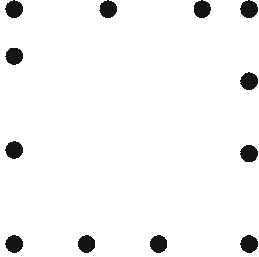}
\caption{Pi-Tag, a relative position based visual fiducial marker.}
\label{pi-tag}
\end{figure}

\begin{figure}[t]
\centering
\begin{subfigure}[b]{0.4\linewidth}
\includegraphics[width=1\linewidth]{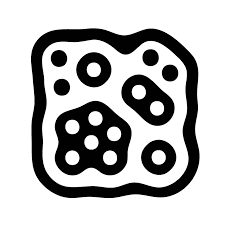}
\caption{ReacTIVision}
\label{reactivision}
\end{subfigure}
\begin{subfigure}[b]{0.4\linewidth}
\includegraphics[width=1\linewidth]{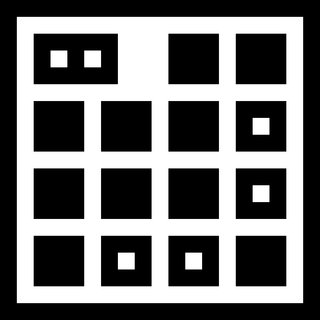}
\caption{TopoTag 4x4}
\label{topotag}
\end{subfigure}
\caption{Two examples of topology based fiducial markers.}
\end{figure}

Relative position based markers encode data through the relative positioning of marker elements to other marker elements, which traditionally are circles.

Pi-Tag\cite{pitag} uses circles arranged with different spacing around the perimeter of a square to encode data. Markers are detected by first detecting circular features, and then validating combinations of these features with projection invariant properties. 

A high rate of false positive rejection can be achieved with this marker design, as the marker elements are localized at a high accuracy and there are strong priors on the locations of the marker elements. However, detection is relatively inefficient, and scales poorly to the amount of detected features. Furthermore, the dictionary size is limited, as the location of markers can only be adjusted in one dimension to maintain the invariants required for detection and decoding. 

\subsection{Topology Based Markers}
Topological patterns provide an alternative to the edge and quad detection step of the square binary markers, or the circle detector step of the relative positioning markers. Topology based markers have uncommon topological structures which can be filtered for by performing hierarchical connected component analysis on the input image after thresholding.

ReacTIVision\cite{reactivision} is a topological based marker system which allows for both detection and decoding purely from topological information. This allows markers to be laid out flexibly, and can decode makers with high distortion.

TopoTag\cite{topotag} improved upon ReacTIVision by encoding the ordering of the data bits using geometric information. This leads to more robust false positive rejection, and higher data density. Furthermore, all points within the marker are used for more accurate pose estimation.

\section{LFTag Design}

\begin{figure}[t]
\centering
\begin{subfigure}[t]{0.3\linewidth}
\includegraphics[width=1\linewidth]{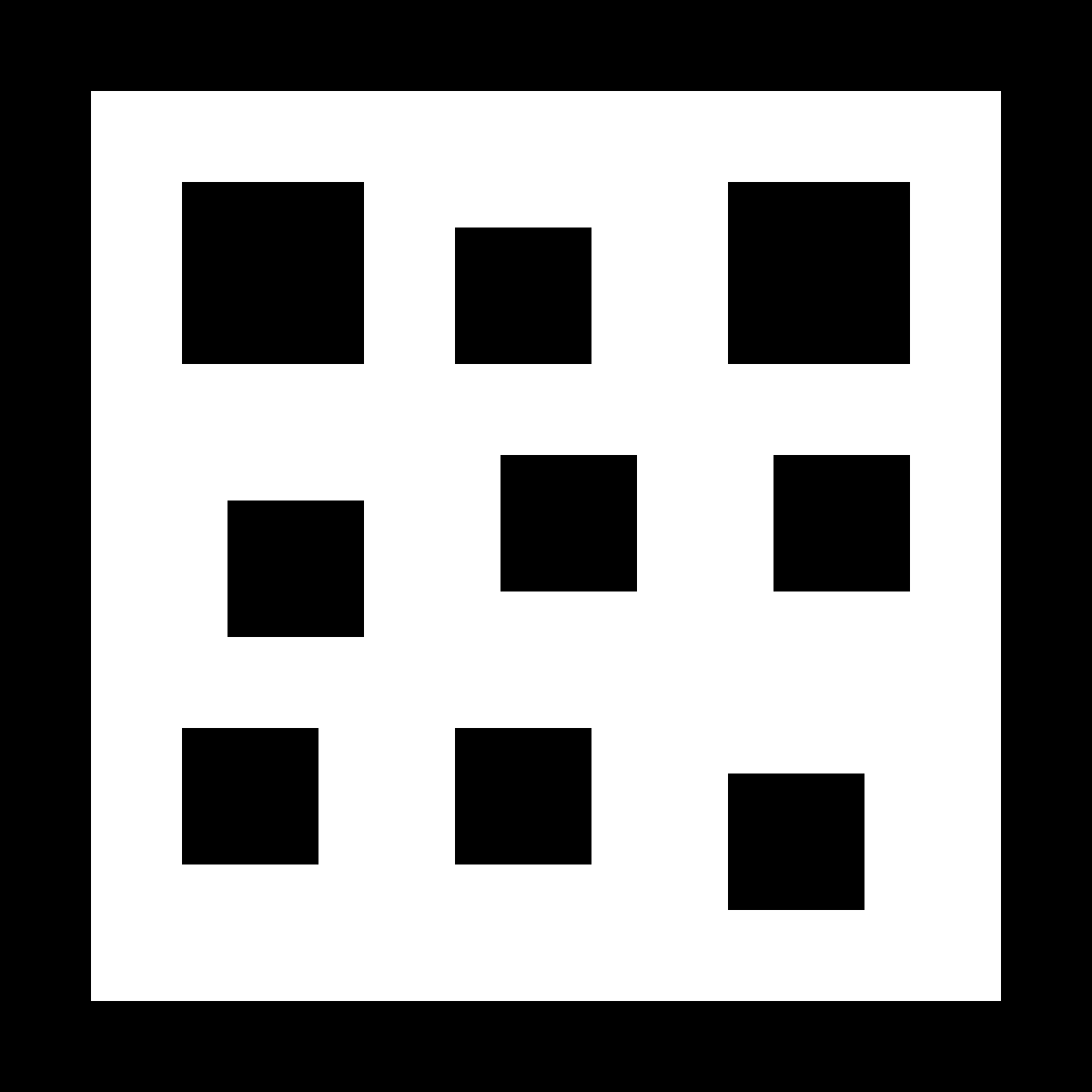}
\caption{}
\end{subfigure}
\begin{subfigure}[t]{0.3\linewidth}
\includegraphics[width=1\linewidth]{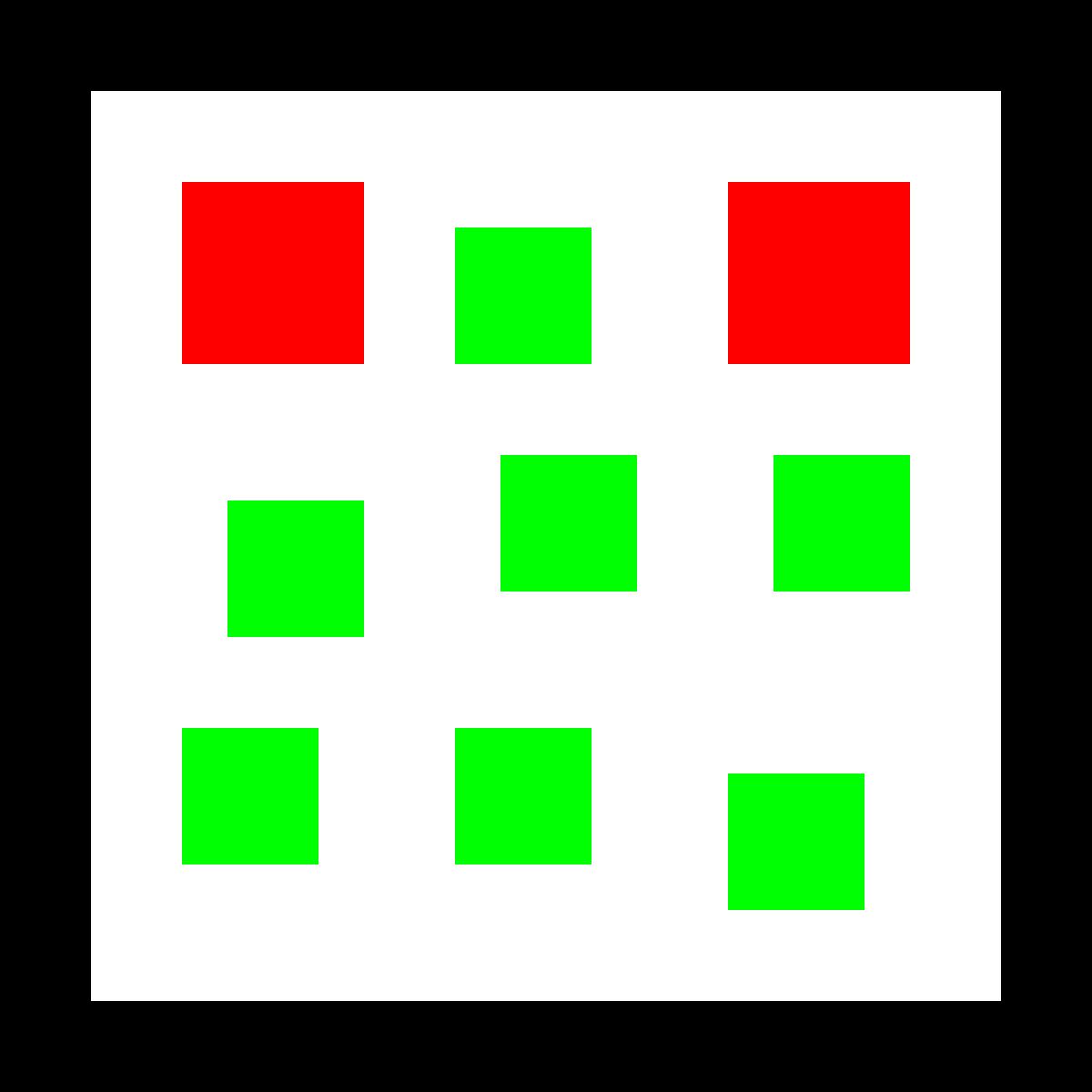}
\caption{}
\end{subfigure}
\begin{subfigure}[t]{0.3\linewidth}
\includegraphics[width=1\linewidth]{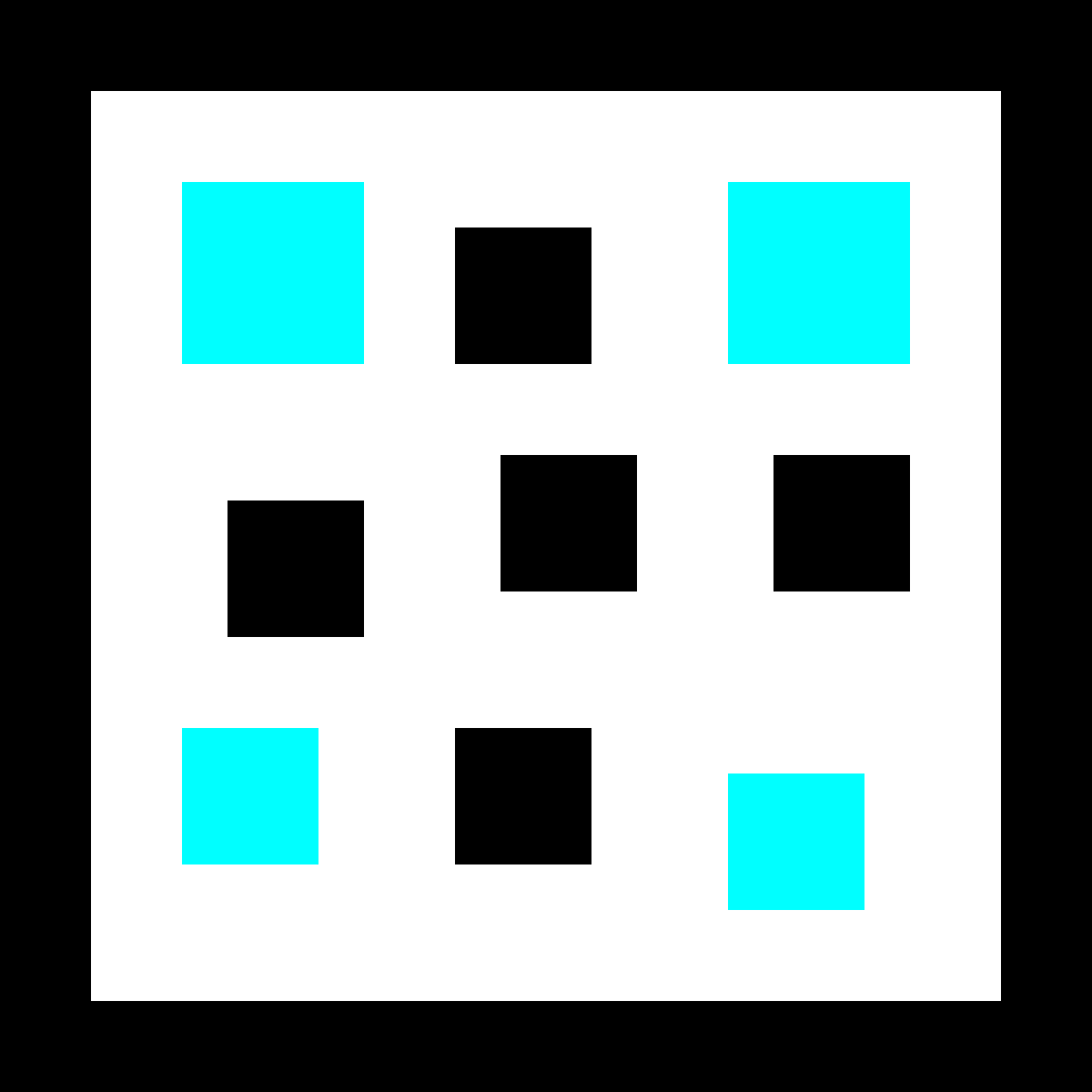}
\caption{}
\end{subfigure}
\caption{A 3x3 LFTag (a) with the baseline regions and data regions shown by red and green respectively in (b), and the key points shown by blue in (c)}
\label{lftag-encoding}
\end{figure}

LFTag is best described as a fiducial marker which performs data encoding with relative positioning of marker elements and detection through topological information. The marker consists of a black border, white background, and \begin{math}n^2\end{math} (the size of a LFTag is denoted by \begin{math}n\end{math}x\begin{math}n\end{math}, e.g. LFTag 3x3 means \begin{math}n = 3\end{math}) black regions inside the marker. Two of the regions (also called "baseline" regions) are larger than the others, and are at opposite sides of the top row of the marker. These points are used to resolve rotational ambiguity. The remaining regions are called "data" regions, and each encode two bits in its relative location. Each data region has 4 possible locations, with the region centroids either shifting up or down, and left or right. The structure is demonstrated graphically in figure~\ref{lftag-encoding}.

By encoding data with subtle shifts within data region positions as opposed to topological structure in TopoTag, LFTag minimizes the high spacial frequency content within the marker that is required for decoding. This allows for low resolving power cameras, which essentially low pass filter the images collected at long range due to either subpar optics, suboptimal focus, or the limited spatial sampling frequency of the sensor (resolution) to still be decoded.

\section{LFTag Detection, Decoding and Localization}
\begin{figure*}%
\centering
\begin{subfigure}[t]{0.5\columnwidth}
\includegraphics[width=1\linewidth]{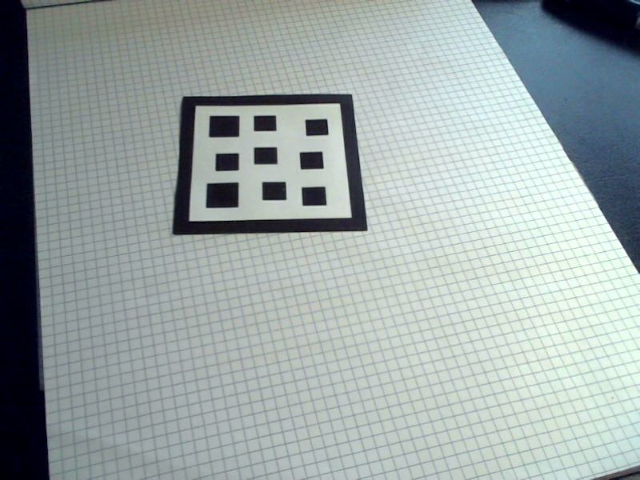}
\caption{Input image}
\end{subfigure}
\begin{subfigure}[t]{0.5\columnwidth}
\includegraphics[width=1\linewidth]{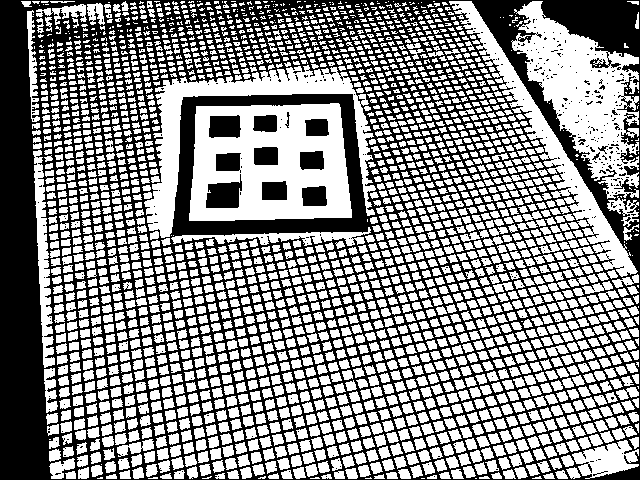}
\caption{Adaptive thresholding}
\end{subfigure}
\begin{subfigure}[t]{0.5\columnwidth}
\includegraphics[width=1\linewidth]{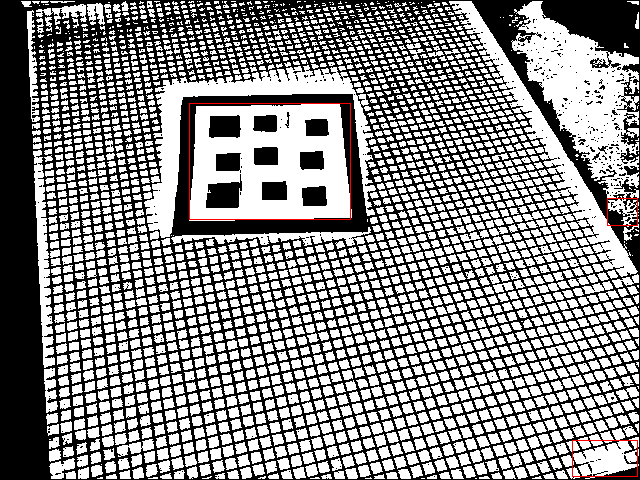}
\caption{Topological filtering}
\end{subfigure}
\begin{subfigure}[t]{0.5\columnwidth}
\includegraphics[width=1\linewidth]{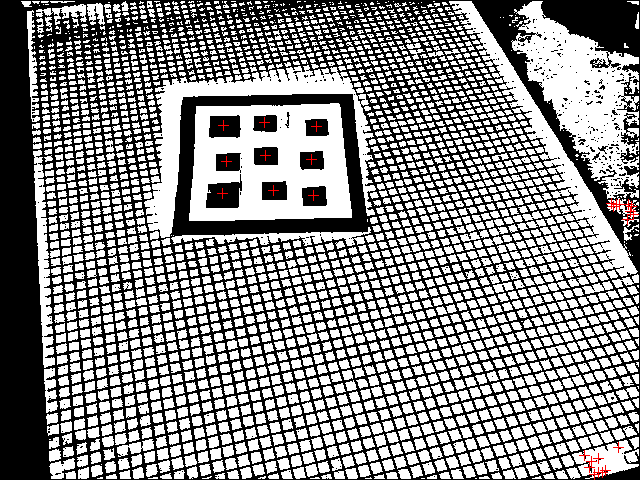}
\caption{Vertex estimation}
\end{subfigure}
\begin{subfigure}[t]{0.5\columnwidth}
\includegraphics[width=1\linewidth]{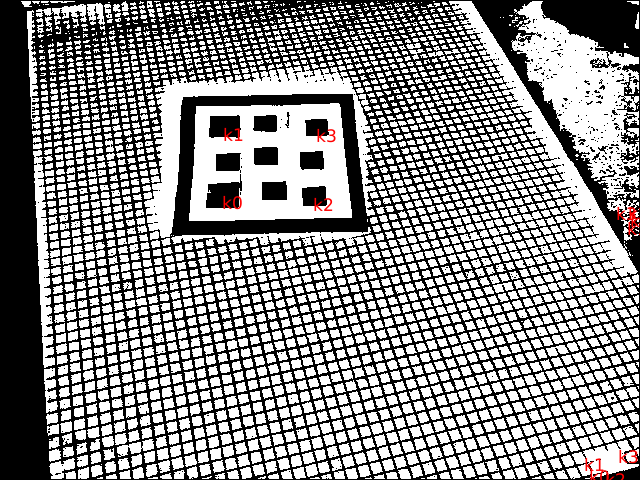}
\caption{Key point identification}
\end{subfigure}
\begin{subfigure}[t]{0.5\columnwidth}
\includegraphics[width=1\linewidth]{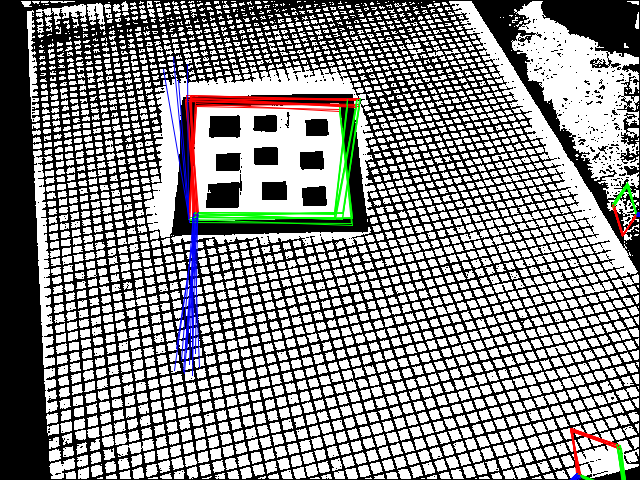}
\caption{Initial poses}
\end{subfigure}
\begin{subfigure}[t]{0.5\columnwidth}
\includegraphics[width=1\linewidth]{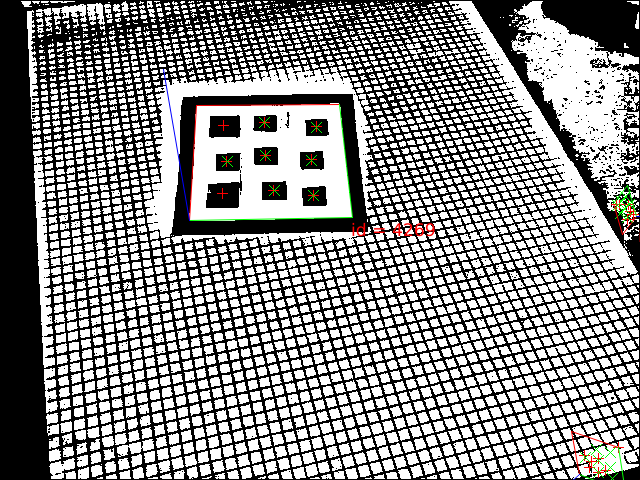}
\caption{Best decoding/initial pose}
\end{subfigure}
\begin{subfigure}[t]{0.5\columnwidth}
\includegraphics[width=1\linewidth]{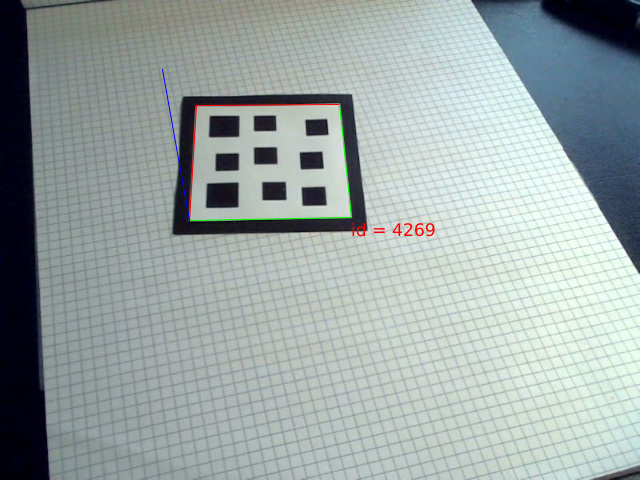}
\caption{Final pose}
\end{subfigure}
\caption{The LFTag detection, decoding and localization pipeline.}
\label{lftag-decoding}
\end{figure*}

The following sections outline the main steps required to detect and localize LFTags within an image. The main steps are also demonstrated with an example in figure~\ref{lftag-decoding}.

\subsection{Binarization}
Binarization is performed using adaptive thresholding, with the threshold map calculated using a square mean window of \begin{math}M * M\end{math} pixels. 

\subsection{Topological Filtering}
After binarization, topological connected component analysis is performed on the image, and a tree structure is produced with the nodes representing regions and the parent-children relationships representing regions that enclose other regions. The connected component analysis is performed using a modified version of the SLCCA algorithm\cite{slcca}, which track hierarchical relationships and performs image moment calculations (for vertex estimation) for all connected components.

The tree structure produced by topological connected component is significantly smaller than the number of pixels in the image, as there often exists large regions of similar brightness in real world images.

This tree structure is filtered to find candidate marker regions, with the specific structure of \begin{math}n^2\end{math} black regions within a white region. Regions with more than \begin{math}n^2\end{math} black regions are also accepted to account for noise, with the largest \begin{math}n^2\end{math} regions used for the next stages of processing.

Furthermore, additional constraints are placed to reduce false positives. If the largest interior region's area is \begin{math}\tau_1\end{math} times the smallest, the candidate is rejected. The marker design only has an factor of 1.77x area difference between the smallest and largest interior region, but the threshold should be set significantly higher to account for perspective distortion, imperfect thresholding, and lighting variation. This is also called the "area constraint".

\subsection{Vertex Estimation}
For each candidate marker found, the centroid of each of the \begin{math}n^2\end{math} interior black regions is found using image moments on an inverted image, after dilation by \begin{math}\delta\end{math} pixels.

Although this is listed as a separate step, region centroids are actually calculated for all regions during the connected component analysis as part of feature vector collection for efficiency.

\subsection{Key Point Identification}
After the candidate marker regions are found, the next step is to identify the four key points located in the corner of the marker, two of them being baseline regions and two of them being data regions in the bottom left and bottom right extent of the marker.

First, the regions are sorted by the inverted zeroth order image moment, after dilation by \begin{math}\delta\end{math} pixels. The two largest regions are picked as the baseline regions, denoted in arbitrary order by \begin{math}B_a\end{math} and \begin{math}B_b\end{math}.

\begin{math}K_0\end{math} and \begin{math}K_1\end{math}
(the top left baseline region and top right baseline region respectively in the canonical marker orientation) corresponds to \begin{math}B_a\end{math} and \begin{math}B_b\end{math}, although in an unknown order.

In order to disambiguate the order of these two regions, first a vector is found from \begin{math}B_a\end{math} to \begin{math}B_b\end{math}. Then, for every region in the marker, a vector is also calculated from \begin{math}B_a\end{math}. The angle between these two vectors is calculated for each region and summed. If the sum of the angles is positive, the \begin{math}B_a\end{math} is \begin{math}K_0\end{math}, otherwise \begin{math}B_b\end{math} is \begin{math}K_0\end{math}. The remaining baseline region is thus \begin{math}K_1\end{math}. This step is performed again, with \begin{math}B_a\end{math} and \begin{math}B_b\end{math} swapped. The region is rejected if the two runs return conflicting results for \begin{math}K_0\end{math}, accounting for false positives with correct topology but incorrect structure. (This is also called the angle constraint)

To locate the remaining two key points, a vector is drawn from \begin{math}K_0\end{math} to \begin{math}K_1\end{math}, and for each remaining region, a vector is found from \begin{math}K_0\end{math}. The list of candidate regions is sorted by this angle, and the list is truncated to  \begin{math}n - 1\end{math}. From this list, the vector that has the longest length is chosen as \begin{math}K_2\end{math}. This process is repeated with \begin{math}K_0\end{math} and \begin{math}K_1\end{math} swapped to find \begin{math}K_3\end{math}.

\subsection{Geometry Filtering}
All data points are checked to ensure that all but \begin{math}n - 2\end{math} data region centroids lie on the same side of the baseline vector. (This is also called the geometry constraint.)

All data and baseline points are also checked to ensure they are not close to being colinear. This is done by performing perpendicular regression, with the region rejected if the residual squared error in pixels is greater than \begin{math}\tau_2\end{math}. (This is also called the collinearity constraint.)

\subsection{Initial Pose Estimation}
To decode the marker, the marker's pose must first be found relative to the camera to locate the data regions within the marker coordinate system. However, as there are only two fixed points within the marker (the centroids of the baseline regions), it is impossible to directly solve for pose at least 4 known marker coordinate to pixel coordinate correspondences as required.

However, as the two data regions which make up the key points each only have 4 possible positions, there is only a total of 16 possible combinations of valid key point locations. This stage produces all 16 possible pose estimations from all combinations of key points using the IPPE\cite{ippe} method to solve the PnP (Perspective-n-Point) problem using the 4 key points.

\subsection{Initial Pose Evaluation}
To test each of the poses estimated in the previous stage, an image gradient based method is used. For each estimated pose, the location of the interior black to white transition is also estimated and projected to back to image space. Along the predicted path, the image gradient is calculated with 3x3 Sobel filters. The magnitude of the image gradient is then averaged over the perimeter length, to produce a metric which evaluates the quality of the particular pose estimation based on the assumed key point locations.

\subsection{Tag Decoding}
For each estimated pose, a decoding is also attempted. For each data region, the 4 possible region centroid positions are transformed into camera coordinates, and compared to the list of region centroids calculated from the image during the vertex estimation stage. The best match is taken for each data region, with the residual squared pixel error accumulated.

For each estimated pose, the residual error from marker decoding is normalized by the square root of the area of the white region in the marker. This is then divided by the pose quality metric calculated in the previous stage to produce a final decoding error metric. The decoding which has the best error metric is then chosen as the final decoding. If the decoding error metric is higher than \begin{math}\tau_3\end{math}, the marker is rejected.

\subsection{Final Pose Estimation}
After the marker has been decoded, the marker coordinates of all region centroids are known. This allows PnP to be solved (using IPPE as above) with all \begin{math}n^2\end{math} points in the marker, which results in a more stable and accurate pose.

\section{LFTag Evaluation}
LFTag is evaluated against AprilTag and TopoTag markers, which represent state of the art in dense binary based and topology based visual fiducial markers respectively. No other markers are evaluated, as AprilTag or TopoTag markers have been shown to perform significantly outperform all other systems in false positive rejection, range or dictionary size to competing markers in \cite{topotag}.

\subsection{Algorithm Setup and Parameters - LFTag}
\begin{itemize}
\item \begin{math}M\end{math} - the height and width of the adaptive thresholding mean window is set at 16 pixels.
\item \begin{math}n\end{math} - the height and width of the marker is tested at both 3 and 4. (Although 2 can be used, it is not tested as it only has a dictionary size of 16 and is significantly less robust, as all points are key points.)
\item \begin{math}\delta\end{math} - the dilation used before calculating region centroids is 1, calculated with \begin{math}L^\infty\end{math} norm.
\item \begin{math}\tau_1\end{math} - the maximum area multiple between the largest and smallest region within a marker before rejection is set at 5.0.
\item \begin{math}\tau_2\end{math} - the minimum perpendicular regression error to reject colinear points is set at 20.0.
\item \begin{math}\tau_3\end{math} - the maximum error metric for marker detection is set at 0.0005 for 3x3 markers and 0.001 for 4x4 markers.
\end{itemize}

The parameters have not been varied for any test.

30 markers are used for evaluation, randomly selected within the dictionary for each family.

\subsection{Algorithm Implementation and Parameters - AprilTag}
The AprilTag 3 detection algorithm was used from the open sourced implementation, with the 16h5, 25h9, 36h11 and 41h12 families evaluated.

All default parameters were used, except for the decimation factor, which was tested at both 2 (default) and 1 (longer range with more compute time) for some experiments.

The first 30 markers of each family were chosen for evaluation (as the markers generation process produces images which are not biased by the data).

\subsection{Algorithm Implementation and Parameters - TopoTag}
The released detector binary\footnote{The TopoTag detector binary is available at \url{https://herohuyongtao.github.io/research/publications/topo-tag/}} was used for TopoTag detection, with the 3x3, 4x4 and 5x5 square markers evaluated.

All default parameters were used, and 30 random markers of each family were chosen for evaluation.

\subsection{False Positives}
False positives are evaluated with the LabelMe dataset\cite{labelme}, which consists of 266,995 images (at time of writing, May 2020) of highly varied indoor and outdoor scenes, which does not contain any fiducial markers. 

The results are shown in table~\ref{false_positive} and an analysis of the different acceptance rates and cumulative false positives at each stage of the LFTag decoding pipeline is provided in table~\ref{false_positive_lftag}.

\begin{table}[htbp]
\caption{False positive results for LFTag, AprilTag and TopoTag on the LabelMe dataset}
\label{false_positive}
\begin{center}
\begin{tabular}{|c|c|c|}
\hline
\textbf{Variant} & \textbf{Dictionary size} & \textbf{False positive detections}\\\hline
\multicolumn{3}{|c|}{AprilTag (No error correction, decimation = 2)} \\ \hline
16h5 & 30 & 2,262 \\ \hline
25h9 & 35 & 3 \\ \hline
36h11 & 587 & 0 \\ \hline
41h12 & 2,115 & 0 \\ \hline

\multicolumn{3}{|c|}{AprilTag (1 bit error correction, decimation = 2)} \\ \hline
16h5 & 30 & 44,577 \\ \hline
25h9 & 35 & 119 \\ \hline
36h11 & 587 & 2 \\ \hline
41h12 & 2,115 & 0 \\ \hline

\multicolumn{3}{|c|}{AprilTag (2 bit error correction, decimation = 2)} \\ \hline
16h5 & 30 & 412,243 \\ \hline
25h9 & 35 & 1,430 \\ \hline
36h11 & 587 & 7 \\ \hline
41h12 & 2,115 & 3 \\ \hline

\multicolumn{3}{|c|}{TopoTag} \\ \hline
3x3 & 128 & 0 \\ \hline
4x4 & 16,384 & 0 \\ \hline
5x5 & 8,388,608 & 0 \\ \hline
\multicolumn{3}{|c|}{LFTag (Proposed)} \\ \hline
3x3 & 16,384 & 1 (see figure~\ref{false_pos_eg}) \\ \hline
4x4 & 268,435,456 & 0 \\ \hline
\end{tabular}
\end{center}
\end{table}

\begin{table}[htbp]
\caption{False positive analysis for LFTag on the LabelMe dataset.}
\label{false_positive_lftag}
\begin{center}
\begin{tabular}{|c|p{2cm}|p{2cm}|}
\hline
\textbf{Detection stage} & \textbf{Stage acceptance} & \textbf{Cumulative false positives}\\\hline
Input images &  & 266,995 \\ \hline
Topological filtering & 443\% & 1,186,431 \\ \hline
Area constraint & 36.8\% & 436,324 \\ \hline
Geometry constraint & 38.6\% & 168,552 \\ \hline
Angle constraint & 99.94\% & 168,452 \\ \hline
Collinearity constraint & 49.4\% & 83,285 \\ \hline
Residual error constraint & 0.0012\% & 1 (see figure~\ref{false_pos_eg}) \\ \hline
\end{tabular}
\end{center}
\end{table}

\begin{figure}[htbp]
\centerline{\includegraphics[width=0.5\linewidth]{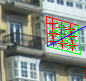}}
\caption{The single false positive detection from the LabelMe dataset, which consists of over 250,000 images.}
\label{false_pos_eg}
\end{figure}

\subsection{Detection Range}
\begin{figure}[t]
\centerline{\includegraphics[width=1\linewidth]{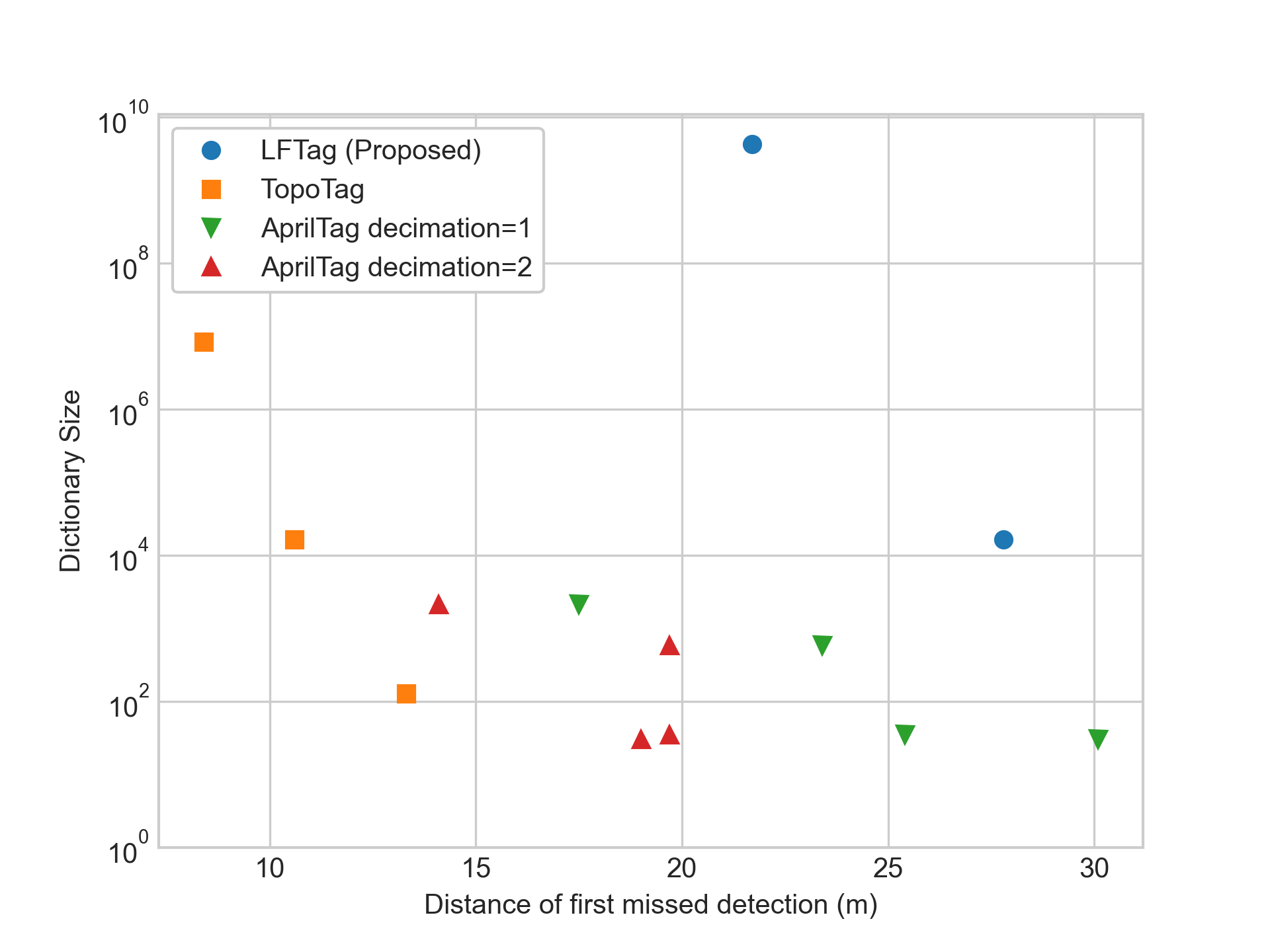}}
\caption{Dictionary size against the distance of the first missed detection.}
\label{range_graph}
\end{figure}

\begin{figure}[htbp]
\centering
\begin{subfigure}[b]{0.45\linewidth}
\includegraphics[width=1\linewidth]{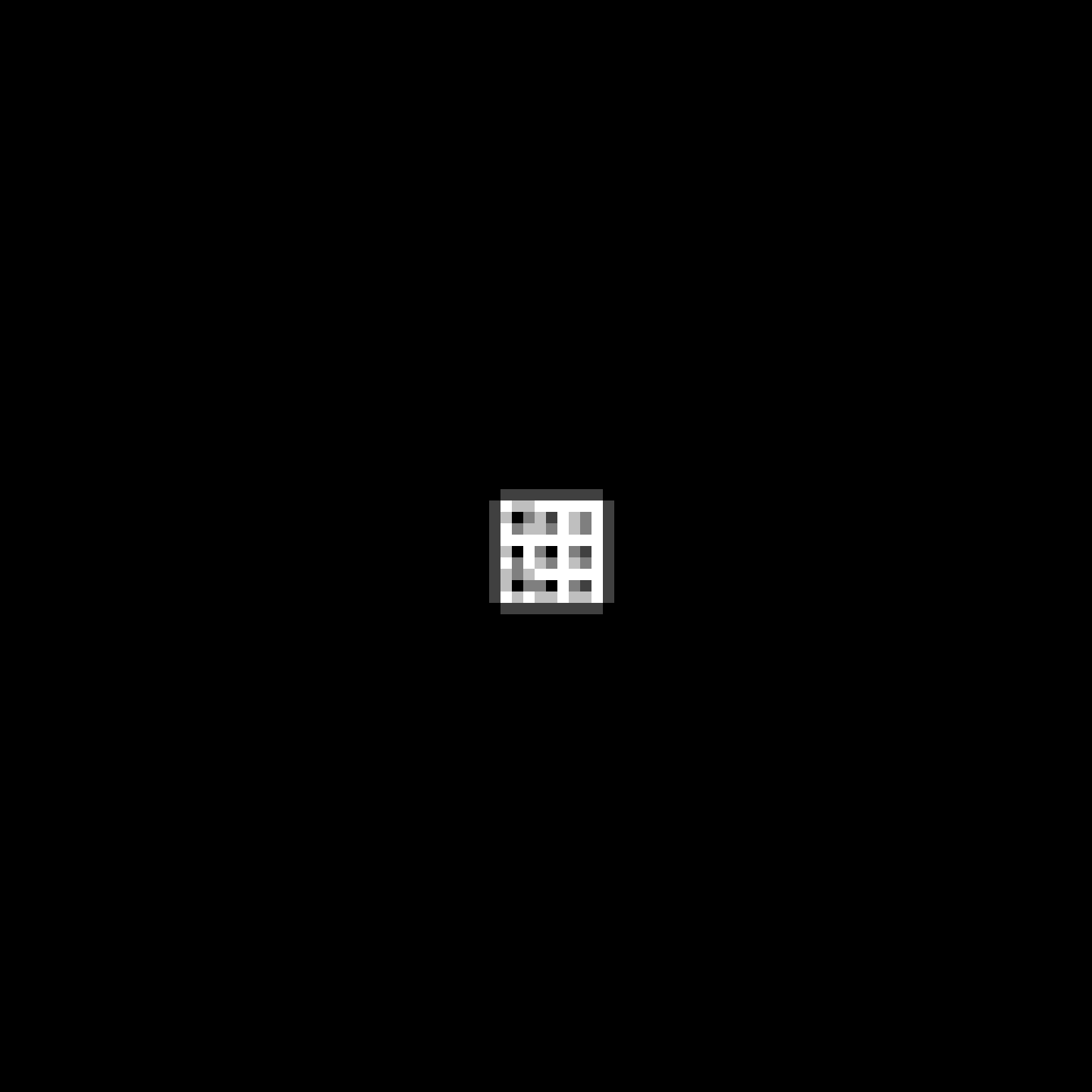}
\end{subfigure}
\begin{subfigure}[b]{0.45\linewidth}
\includegraphics[width=1\linewidth]{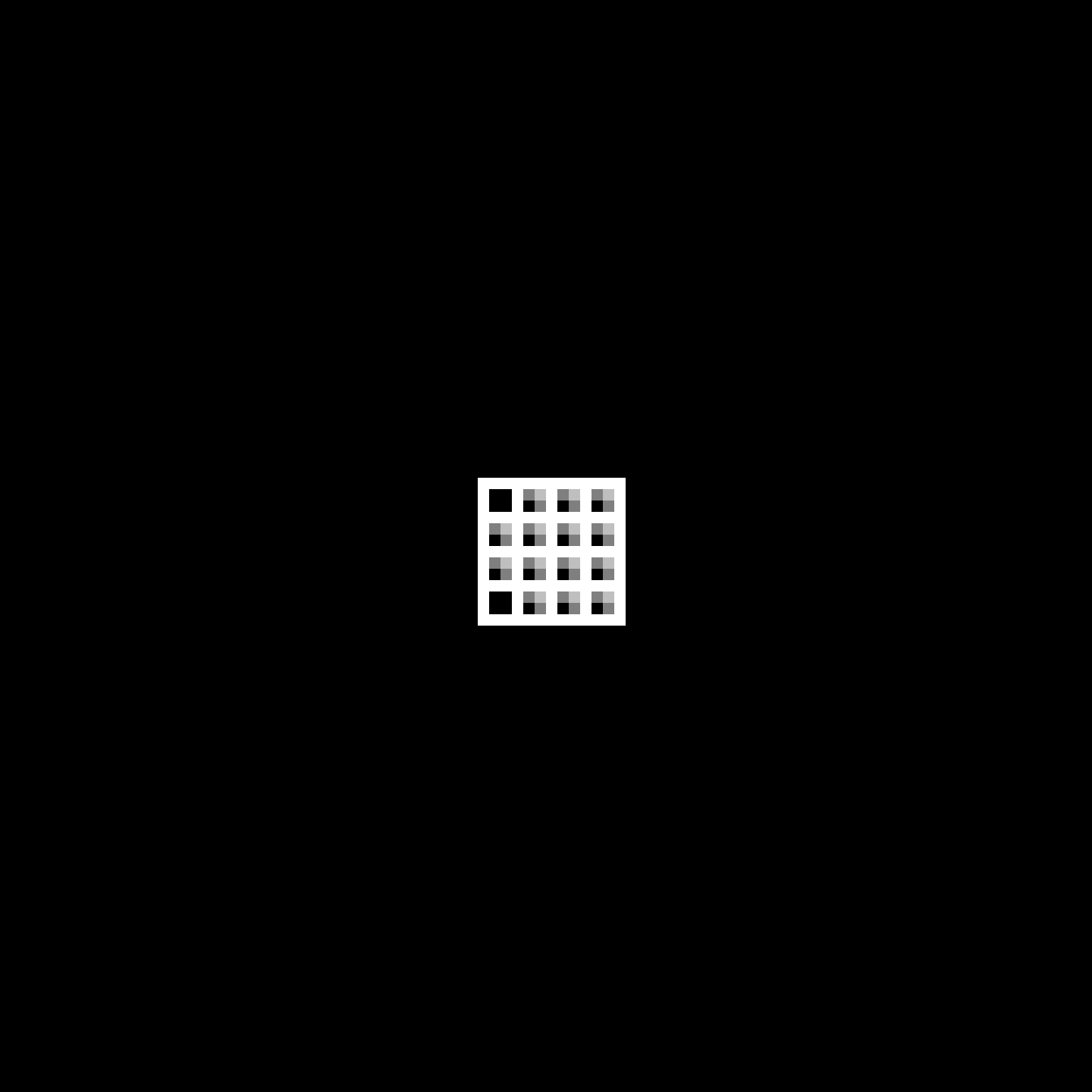}
\end{subfigure}
\caption{Input images for LFTag detections at maximum range.}
\label{max_range}
\end{figure}

Detection range is evaluated in a simulated environment. A virtual camera renders 640x480 anti-aliased images of markers, normalized to a one meter side length, with a focal length of 320 pixels. 30 markers are used for each marker type, with correct detection probability recorded at each distance. The range at the first missed detection and 20\% missed detection probability threshold is recorded in table~\ref{range}. The dictionary size is also plotted against the distance of the first missed detection, in figure~\ref{range_graph}. The images for the last successful detection is also shown in figure~\ref{max_range}.

\begin{figure}[t]
\centering
\begin{subfigure}[b]{0.45\linewidth}
\includegraphics[width=1\linewidth]{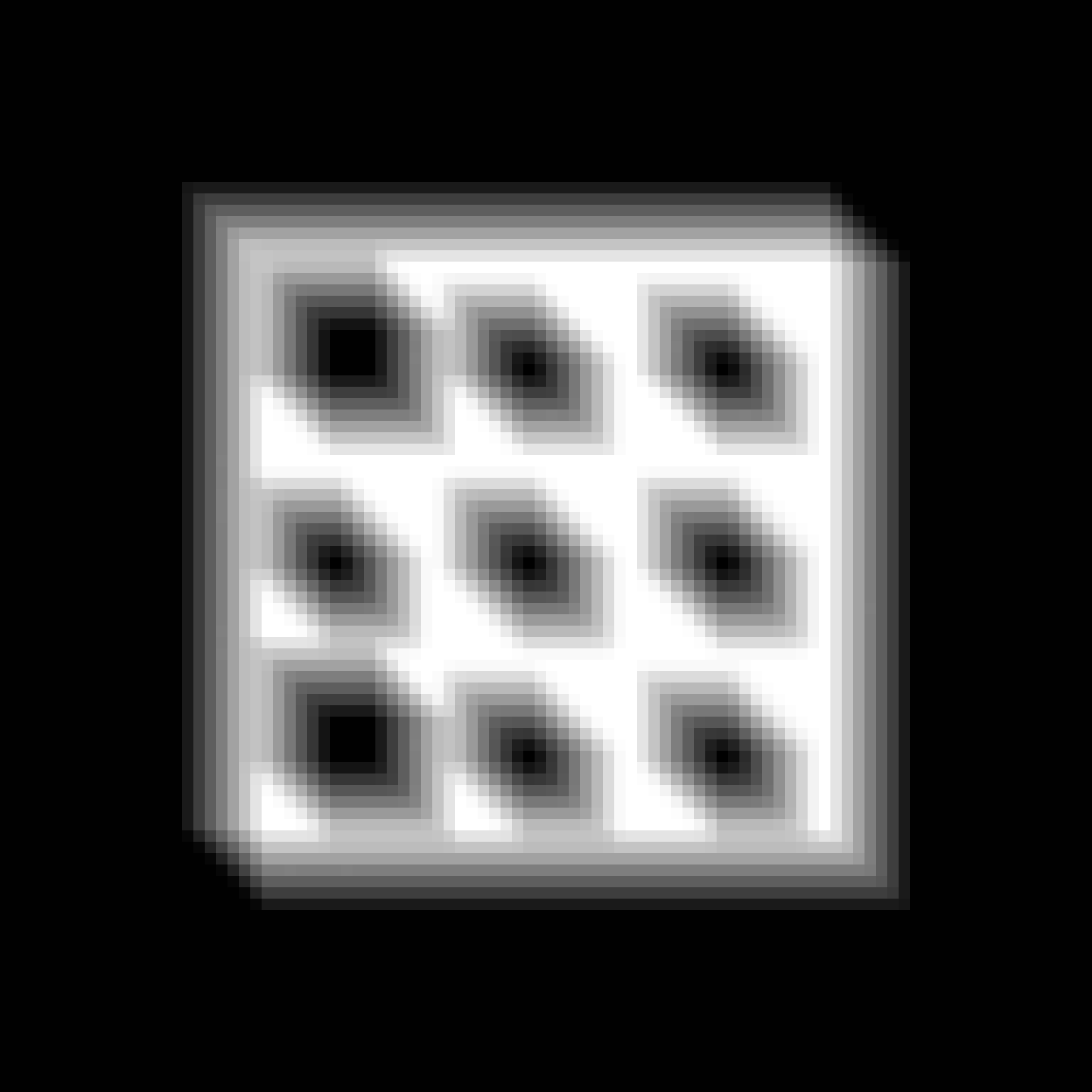}
\end{subfigure}
\begin{subfigure}[b]{0.45\linewidth}
\includegraphics[width=1\linewidth]{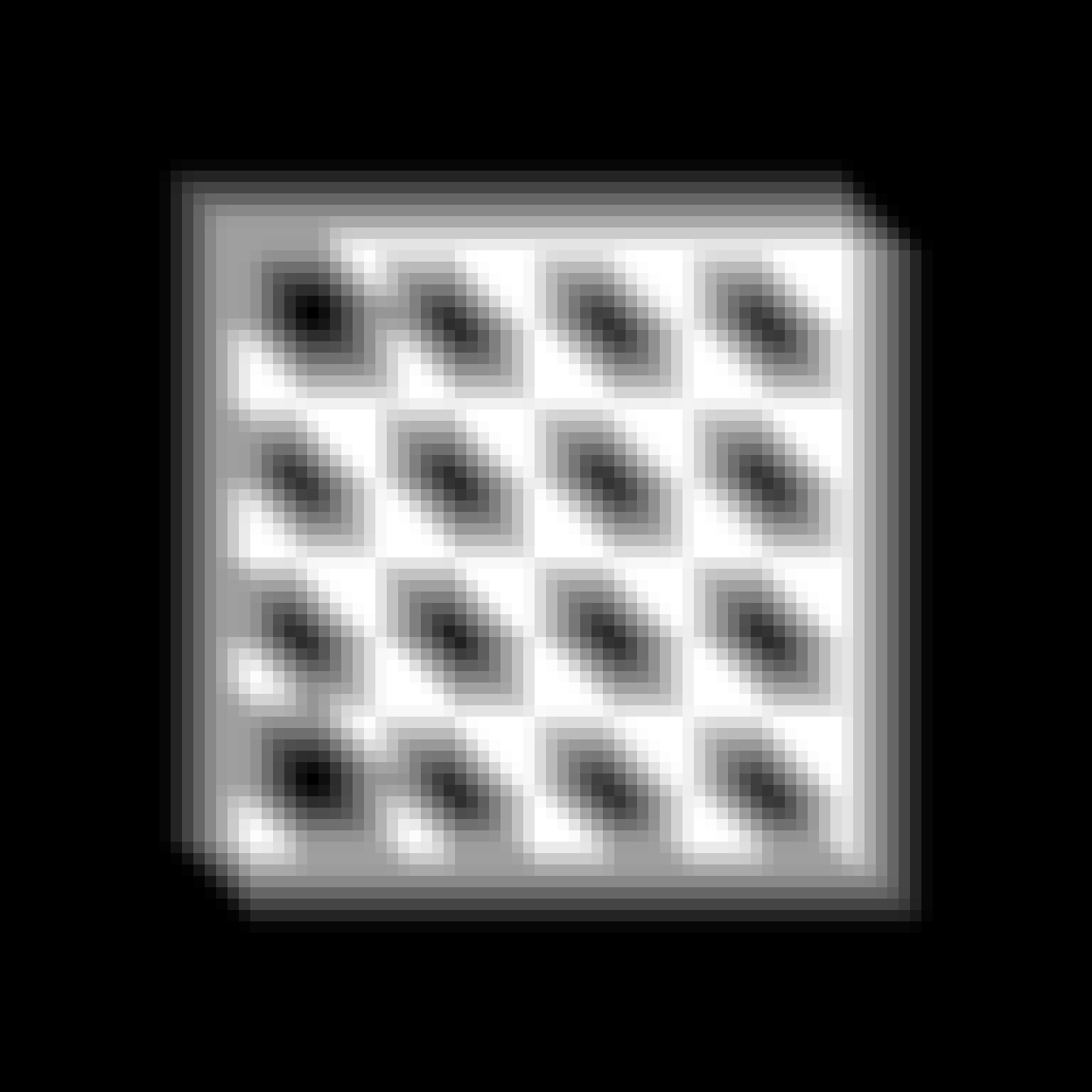}
\end{subfigure}
\caption{Input images for LFTag detections at maximum motion blur.}
\label{max_blur}
\end{figure}

\begin{table}[htbp]
\caption{Detection range results for LFTag, AprilTag and TopoTag on the synthetic dataset. }
\label{range}
\begin{center}
\begin{tabular}{|c|c|c|c|}
\hline
\textbf{Variant} & \textbf{Dictionary size} & \textbf{First missed (m)} & \textbf{20\% missed (m)}\\\hline
\multicolumn{4}{|c|}{AprilTag (2 bit error correction, decimation = 1)} \\ \hline
16h5 & 30 & 30.1 & 33.0 \\ \hline
25h9 & 35 & 25.4 & 27.1 \\ \hline
36h11 & 587 & 23.4 & 23.9 \\ \hline
41h12 & 2,115 & 17.5 & 21.7 \\ \hline
\multicolumn{4}{|c|}{AprilTag (2 bit correction, decimation = 2)} \\ \hline
16h5 & 30 & 19.0 & 19.0 \\ \hline
25h9 & 35 & 19.7 & 19.7 \\ \hline
36h11 & 587 & 19.7 & 19.7 \\ \hline
41h12 & 2,115 & 14.1 & 14.3 \\ \hline
\multicolumn{4}{|c|}{TopoTag} \\ \hline
3x3 & 128 & 13.3 & 13.3 \\ \hline
4x4 & 16,384 & 10.6 & 10.6 \\ \hline
5x5 & 8,388,608 & 8.39 & 8.39 \\ \hline
\multicolumn{4}{|c|}{LFTag (Proposed)} \\ \hline
3x3 & 16,384 & 27.8 & 27.8 \\ \hline
4x4 & 268,435,456 & 21.7 & 21.7 \\ \hline
\end{tabular}
\end{center}
\end{table}

\subsection{Motion Blur}
\begin{figure}[htbp]
\centerline{\includegraphics[width=1\linewidth]{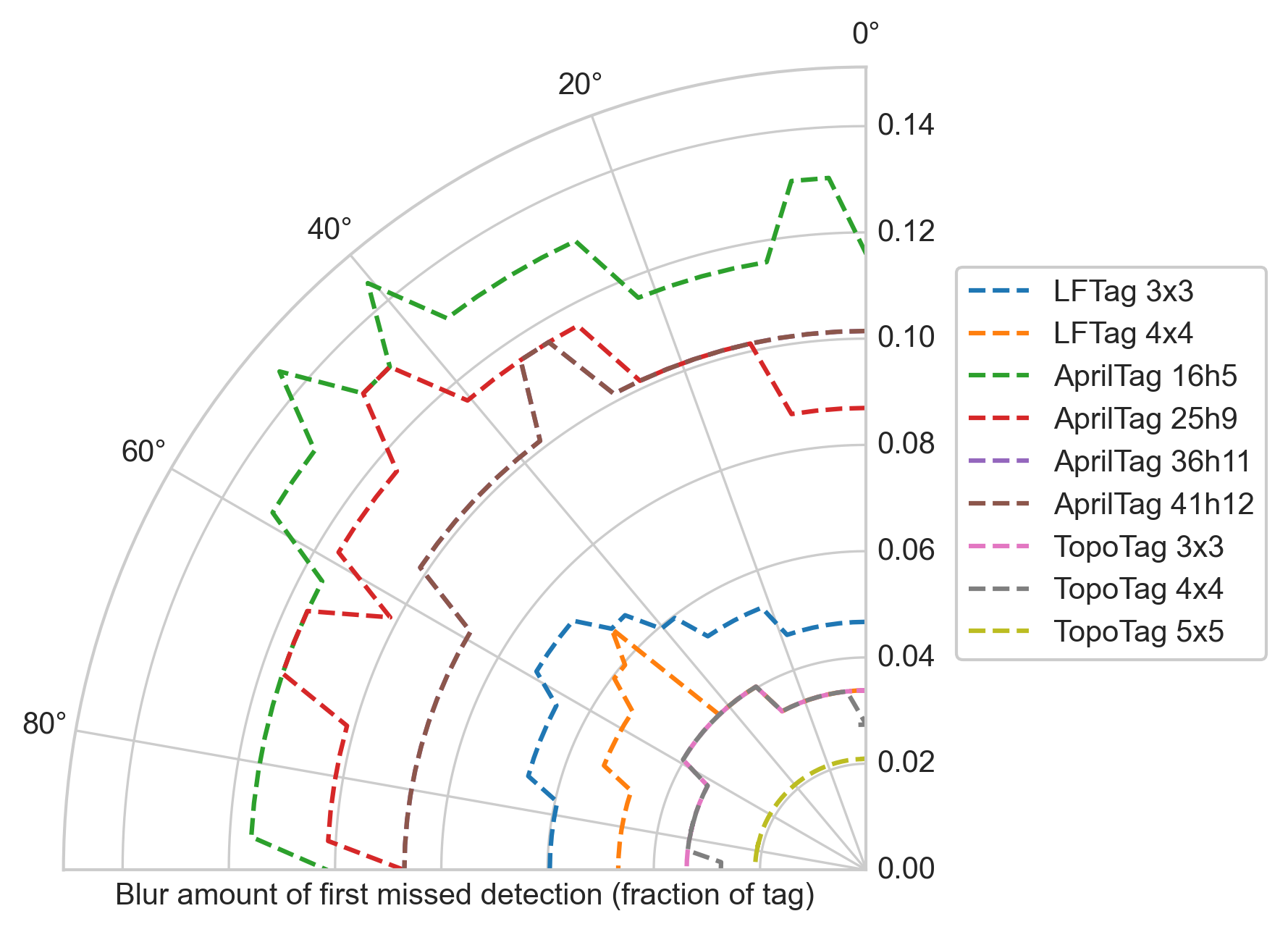}}
\caption{Angle and magnitude of the first missed detection due to motion blur.}
\label{blur_polar}
\end{figure}

\begin{figure}[htbp]
\centering
\begin{subfigure}[b]{0.45\linewidth}
\includegraphics[width=1\linewidth]{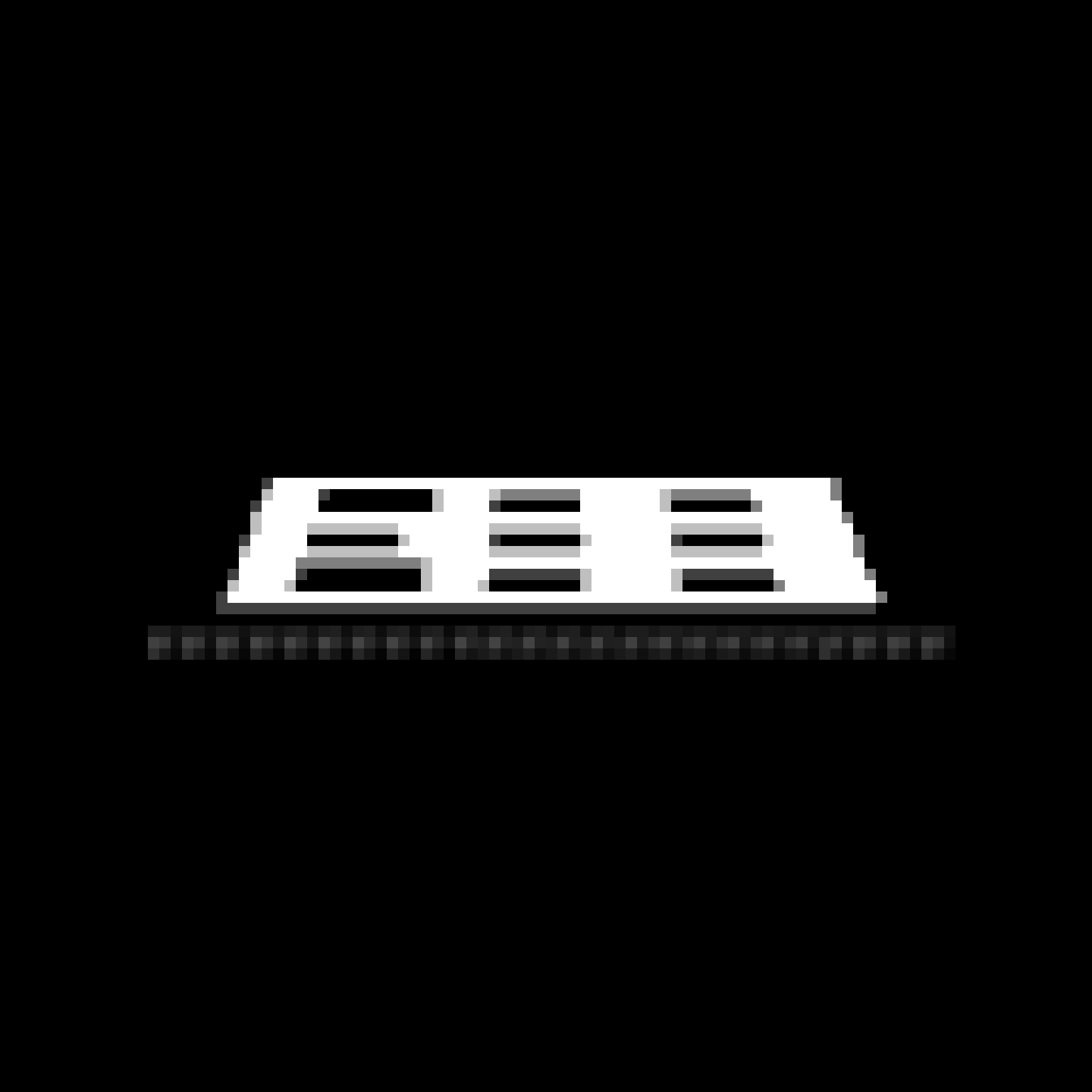}
\end{subfigure}
\begin{subfigure}[b]{0.45\linewidth}
\includegraphics[width=1\linewidth]{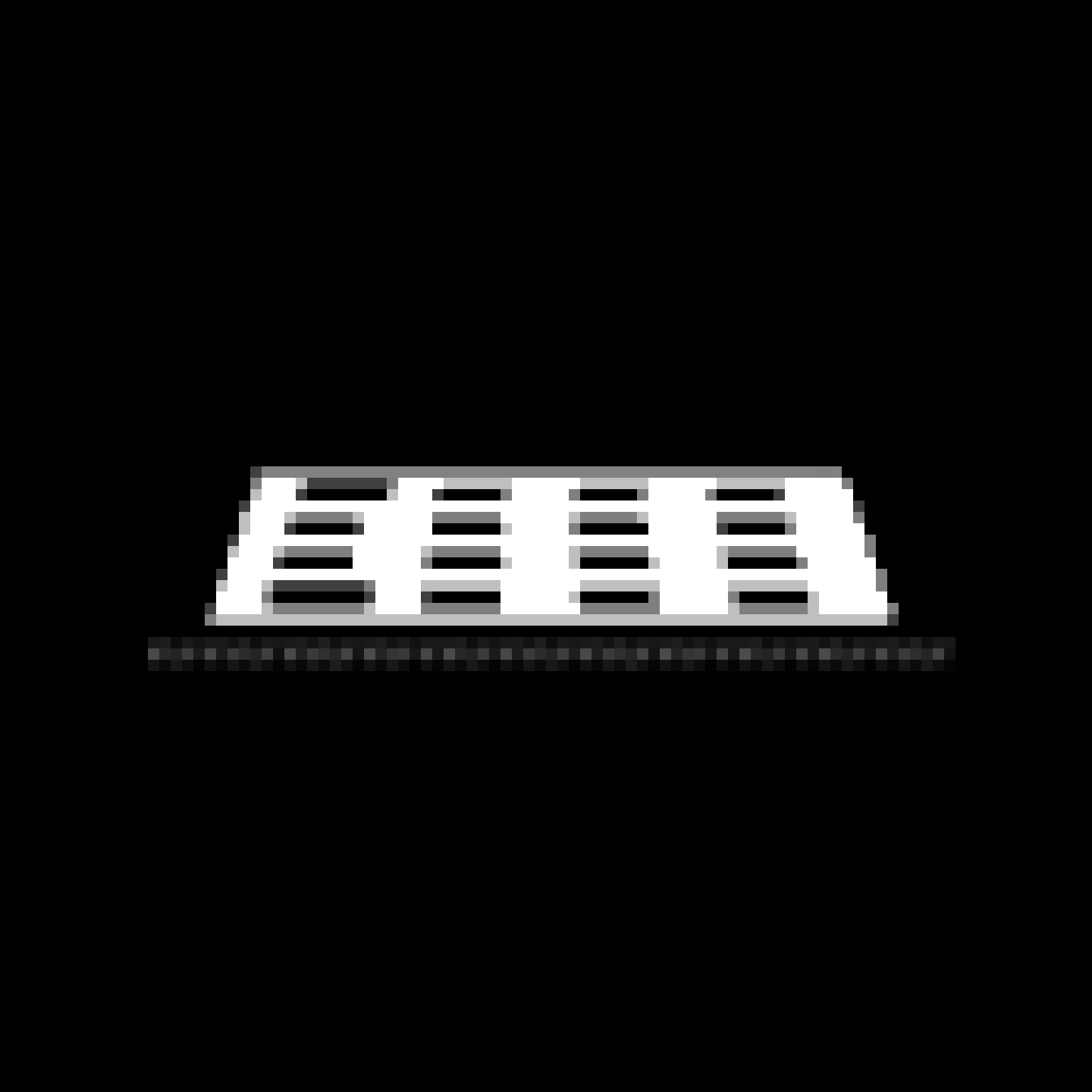}
\end{subfigure}
\caption{Input images for LFTag detections at maximum angle.}
\label{max_angle}
\end{figure}

\begin{table}[htbp]
\caption{Angle results for LFTag, AprilTag and TopoTag on the synthetic dataset.}
\label{angle}
\begin{center}
\begin{tabular}{|c|c|c|}
\hline
\textbf{Variant} & \textbf{Dictionary size} & \textbf{First missed detection (degree)}\\\hline
\multicolumn{3}{|c|}{AprilTag (2 bit error correction, decimation = 1)} \\ \hline
16h5 & 30 & 77.5\degree \\ \hline
25h9 & 35 & 77.5\degree \\ \hline
36h11 & 587 & 77.5\degree \\ \hline
41h12 & 2,115 & 77.5\degree \\ \hline
\multicolumn{3}{|c|}{AprilTag (2 bit correction, decimation = 2)} \\ \hline
16h5 & 30 & 74.5\degree \\ \hline
25h9 & 35 & 76\degree \\ \hline
36h11 & 587 & 74.5\degree \\ \hline
41h12 & 2,115 & 76\degree \\ \hline
\multicolumn{3}{|c|}{TopoTag} \\ \hline
3x3 & 128 & 61.5\degree \\ \hline
4x4 & 16,384 & 53\degree \\ \hline
5x5 & 8,388,608 & 45\degree \\ \hline
\multicolumn{3}{|c|}{LFTag (Proposed)} \\ \hline
3x3 & 16,384 & 77.5\degree \\ \hline
4x4 & 268,435,456 & 76\degree \\ \hline
\end{tabular}
\end{center}
\end{table}

Motion blur is a significant challenge for fiducial markers, and is often the cause for missed detections in the real world, especially in applications with high rotation or translation rates, and/or insufficient lighting. This is evaluated with a dataset generated by using a uniform linear point spread function of variable length and angle applied to marker images at constant range to generate synthetic motion blur.

The blur is varied between 0\degree (vertical) to 90\degree (horizontal), and the magnitude of the blur varies from 0 to 15\% the edge length of the marker. 30 markers are used for each marker type, with correct detection probability recorded at each angle and magnitude.

The first missed detection for all markers at all angles is shown graphically in figure~\ref{blur_polar}. The image for the last successful detection is shown in figure~\ref{max_blur}.

\subsection{Marker Angle}

Detection probability is also tested at grazing angles. 30 markers are used for each marker type, with correct detection probability recorded at each angle and magnitude. The angle of the first missed detection is shown in table~\ref{angle}. The image for the last successful detection is shown in figure~\ref{max_angle}.

\subsection{Detector Performance}
Performance is evaluated on a desktop computer, with an AMD Ryzen 7 2700X CPU (8 cores, 16 threads @ 4.30GHz) and 32GB of RAM. All algorithms are run on a single thread.

The LFTag detector is implemented in Rust, and the OpenCV implementation of the IPPE PnP algorithm was used. The detector is single threaded and has not been performance optimized.

Evaluations are performed with 1280x720 images, collected from a webcam where one fiducial marker per type is visible. Input parsing and decompression time is not measured, and pose estimation is performed for all detected markers for all implementations.

\begin{table}[htbp]
\caption{Performance results for LFTag, AprilTag and TopoTag.}
\label{perf}
\begin{center}
\begin{tabular}{|c|c|}
\hline
\textbf{Marker} & \textbf{Runtime (ms)}\\\hline
Apriltag (Decimate=1) & 32 \\ \hline
Apriltag (Decimate=2) & 10 \\ \hline
TopoTag\footnotemark[1] & 20 \\ \hline
LFTag (Proposed) & 21 \\ \hline
\end{tabular}
\end{center}
\end{table}

\footnotetext[1]{As the released binary does not provide timing information, performance is evaluated with the author's reproduction of the algorithm in Rust, available at \url{https://github.com/kingoflolz/topotag}}

\begin{table}[htbp]
\caption{Performance breakdown for the LFTag detector on 1280x720 images.}
\label{perf_breakdown}
\begin{center}
\begin{tabular}{|c|c|c|}
\hline
\textbf{Stage} & \textbf{Time (us)} & \textbf{Share of total}\\\hline
Adaptive Thresholding & 489 & 2.3\% \\ \hline
Topological filtering & 18,285 & 87\% \\ \hline
Key point identification & 562 & 2.7\% \\ \hline
Geometric filtering & 118 & 0.5\% \\ \hline
Initial pose estimation & 1,320 & 6.3\% \\ \hline
Decoding & 112 & 0.5\% \\ \hline
Final pose estimation & 180 & 0.9\% \\ \hline

\end{tabular}
\end{center}
\end{table}

\section{Discussion}

The most prominent result from this paper is the novel detection and encoding scheme. The use of topological filtering produces roughly an order of magnitude less candidate regions compared to state of the art square marker detectors (1.2 million regions with LFTag compared to 9 million quads with the AprilTag 3 detector on the LabelMe dataset). Furthermore, strong geometric filtering schemes means only 7\% of these candidate regions attempt to be decoded\footnote{The geometric filtering stages only has a limited effect in reducing false positives, as almost all regions that is rejected by the geometric filtering will also be rejected by the residual error constraints. However, as the speed of geometric filtering stage is very fast, it reduces the number of times the relatively expensive initial pose estimation stage will be called.}. The highly accurate vertex localization and strong priors of the locations of the regions inside the marker ensure that all but one false positive is rejected in over 250 thousand images, without sacrificing any data bits for false positive rejection.

Although TopoTag can achieve very low false positive rates without sacrificing any marker bits, the data encoded is fragile to low pass filtering, as the complex topological structure requires multiple nested levels opposite colors. This leads to limited range and robustness compared to LFTag.

Furthermore, the relative position based encoding scheme encodes a large amount of data by efficiently utilizing the limited spatial bandwidth when detecting marker from long range. When combined with the inherent topological and geometric false positive resistance of the marker, all marker bits can be used for data encoding. This leads to a 2-5 order of magnitude improvement in dictionary sizes at a given range and false positive rate.

The robustness of the proposed fiducial markers is also evaluated, and demonstrates usable performance for applications in the real world. Square binary markers perform significantly better than the proposed marker design and TopoTag, as the topological information is relatively easier to disrupt than the edge features used to locate square markers.

The performance of LFTag at grazing angles matches the performance of square binary markers, as the low spatial frequency of the marker enables the markers to still be picked up with very limited vertical resolution. 

Although LFTag does not achieve state-of-the-art robustness, it is still usable in the real world, especially considering its significantly higher encoding density.

LFTag is also highly scalable. Although the largest marker tested is only 4x4 in size, the implementation has been tested to work up to at least 8x8. The dense binary encoding allows for large amounts of data (over 100 bits) to be stored per marker, while requiring no expensive dictionary generation step.

However, there are several limitations to the proposed marker design; occlusion and bending of markers.

Due to the topological detection method of the marker, a small amount of occlusion can change the topological structure and prevent detection. Furthermore, occlusion of any key point, even if the topology of the marker is otherwise maintained, means the marker cannot be decoded correctly.

As the marker design relies on accurate priors of the centroids of the marker regions from localizing the key points, the marker is sensitive to bending. When a bent marker attempts to be detected, it is possible the wrong data is read if the region centroids of the bent marker approximates another valid configuration of a flat marker, or the marker detection can be missed if the residual error is too high.

\section{Future Work}
There are several interesting aspects of the proposed design which can be improved: false positive and speed with fixed key points, using closed form solutions for initial pose estimation, bent marker resistance by solving for distortion in decoding, error correction, and higher data density with alternative marker structures.

One of the slower stages of marker detection is initial pose estimation, as 16 candidate poses need to be solved for each marker. This can possibly be improved with a small change to the system so one or both of the data carrying key points stay fixed. If one data carrying key point is fixed, a 4x improvement is expected in both the speed of initial pose estimation, and also in false positive rates. This increases to 16x if both data carrying key points are fixed. However, this comes with a corresponding 4x and 16x reduction in dictionary sizes. Conversely, if computation speed and false positive detections are not an issue, one of the baseline regions can be modified to be a data region, to achieve 4x larger dictionary sizes.

The IPPE PnP algorithm can be solved for analytically, given known marker points. This has been done for squares in \cite{ippe}, and results in a significant speed improvement for pose estimation of square markers. It should also be possible for analytical solutions of all 16 possible key point locations to be calculated, which should result in a corresponding speedup in the initial pose estimation of the LFTag detector algorithm.

Another potential improvement to solve the robustness of markers in bending is to directly solve for the amount of bending during either initial pose estimation from the white to black inner border, or during decoding from residual error. In most cases, bending can be modelled with a two variable model of bend radius and angle, which should significantly improve robustness to bending. However, care needs to be taken to ensure false positive rates remain acceptable, as this weakens the priors on the expected region locations.

The proposed marker currently does not provide any error detection or correction, using all marker bits for data. This is still highly robust as data decoding is based on region centroids, which is very noise resistant. However, for many applications, an even lower false positive rate or partial occlusion resistance is more important, and the current dictionary size too large. The dense binary data encoding is well suited to efficient error correction schemes such as Hamming codes or Reed-Solomon error correction. Furthermore, as there are strong geometric priors for a correctly decoded markers, many errors can be treated as erasures instead, which allows for more bits to be corrected.

Additionally, a square or grid structure is likely not the most optimal arrangement for the data regions to maximize the utilization of spatial bandwidth. This is seen in the motion blur test, where for almost all markers, and expecially for LFTag 3x3, the 45 degree axis is more resistant to blurring. Perhaps a hexagonal arrangement can be used to optimize the packing density of the data regions. Furthermore, with a hexagonal arrangement of data regions, each data region can likely encode 6 different possible locations, leading to a higher data density.

This paper only focused on the detection performance of the various fiducial markers, and did not evaluate the quality of the pose estimations. In general, the better the vertex estimation and number of marker-camera correspondences, the lower the pose error and jitter. Combined with empirical evaluations, it is expected that LFTag performs well in this benchmark, but an objective evaluation would be valuable. 

\section{Conclusions}
This paper presents LFTag, a visual fiducial system based on topological detection and relative positioning data encoding. A very low false positive rate was achieved with highly accurate, sub-pixel localization of interior regions combined with strong priors due to the marker construction. The construction of the marker also resolves rotational ambiguity, which combined with the low false positive rate allows for all marker bits to be used for data.

As the relative positioning based data encoding maximizes the use of spatial bandwidth and all marker bits are used for data, significant advancements are made in the dictionary size to detection range tradeoffs of existing fiducial markers, offering 546 times the dictionary size of AprilTag 25h9 with LFTag 3x3 or 126 thousand times the dictionary size of AprilTag 41h12 with LFTag 4x4 while simultaneously achieving longer detection rage, or more than twice the detection range of TopoTag 4x4 at the same dictionary size with LFTag 3x3.
\bibliographystyle{IEEEtran} 
\bibliography{lftag} 

\begin{thebibliography}{10}
\providecommand{\url}[1]{#1}
\csname url@samestyle\endcsname
\providecommand{\newblock}{\relax}
\providecommand{\bibinfo}[2]{#2}
\providecommand{\BIBentrySTDinterwordspacing}{\spaceskip=0pt\relax}
\providecommand{\BIBentryALTinterwordstretchfactor}{4}
\providecommand{\BIBentryALTinterwordspacing}{\spaceskip=\fontdimen2\font plus
\BIBentryALTinterwordstretchfactor\fontdimen3\font minus
  \fontdimen4\font\relax}
\providecommand{\BIBforeignlanguage}[2]{{%
\expandafter\ifx\csname l@#1\endcsname\relax
\typeout{** WARNING: IEEEtran.bst: No hyphenation pattern has been}%
\typeout{** loaded for the language `#1'. Using the pattern for}%
\typeout{** the default language instead.}%
\else
\language=\csname l@#1\endcsname
\fi
#2}}
\providecommand{\BIBdecl}{\relax}
\BIBdecl

\bibitem{1544677}
L.~{Naimark} and E.~{Foxlin}, ``Encoded led system for optical trackers,'' in
  \emph{Fourth IEEE and ACM International Symposium on Mixed and Augmented
  Reality (ISMAR'05)}, 2005, pp. 150--153.

\bibitem{huang2019lidartag}
J.-K. Huang, M.~Ghaffari, R.~Hartley, L.~Gan, R.~M. Eustice, and J.~W. Grizzle,
  ``Lidartag: A real-time fiducial tag using point clouds,'' 2019.

\bibitem{10.1145/1531326.1531404}
\BIBentryALTinterwordspacing
A.~Mohan, G.~Woo, S.~Hiura, Q.~Smithwick, and R.~Raskar, ``Bokode:
  Imperceptible visual tags for camera based interaction from a distance,''
  \emph{ACM Trans. Graph.}, vol.~28, no.~3, Jul. 2009. [Online]. Available:
  \url{https://doi.org/10.1145/1531326.1531404}
\BIBentrySTDinterwordspacing

\bibitem{artag}
M.~Fiala, ``Artag, an improved marker system based on artoolkit,'' 01 2004.

\bibitem{apriltag}
E.~Olson, ``{AprilTag}: A robust and flexible visual fiducial system,'' in
  \emph{Proceedings of the {IEEE} International Conference on Robotics and
  Automation ({ICRA})}.\hskip 1em plus 0.5em minus 0.4em\relax IEEE, May 2011,
  pp. 3400--3407.

\bibitem{apriltag3}
M.~Krogius, A.~Haggenmiller, and E.~Olson, ``Flexible layouts for fiducial
  tags,'' in \emph{Proceedings of the {IEEE/RSJ} International Conference on
  Intelligent Robots and Systems {(IROS)}}, 2019.

\bibitem{pitag}
F.~Bergamasco, A.~Albarelli, and A.~Torsello, ``Pi-tag: A fast image-space
  marker design based on projective invariants,'' \emph{Machine Vision and
  Applications}, vol.~24, 08 2013.

\bibitem{reactivision}
M.~Kaltenbrunner and R.~Bencina, ``reactivision: a computer-vision framework
  for table-based tangible interaction,'' 01 2007.

\bibitem{topotag}
G.~Yu, Y.~Hu, and J.~Dai, ``Topotag: A robust and scalable topological fiducial
  marker system,'' \emph{IEEE Transactions on Visualization and Computer
  Graphics}, vol.~PP, pp. 1--1, 04 2020.

\bibitem{slcca}
M.~Klaiber, ``A parallel and resource-efficient single lookup connected
  components analysis architecture for reconfigurable hardware,'' Ph.D.
  dissertation, 07 2016.

\bibitem{ippe}
T.~Collins and A.~Bartoli, ``Infinitesimal plane-based pose estimation,''
  \emph{International Journal of Computer Vision}, vol. 109, 09 2014.

\bibitem{labelme}
\BIBentryALTinterwordspacing
B.~C. Russell, A.~Torralba, K.~P. Murphy, and W.~T. Freeman, ``Labelme: A
  database and web-based tool for image annotation,'' \emph{Int. J. Comput.
  Vision}, vol.~77, no. 1–3, p. 157–173, May 2008. [Online]. Available:
  \url{https://doi.org/10.1007/s11263-007-0090-8}
\BIBentrySTDinterwordspacing

\end{thebibliography}
\end{document}